\documentclass[runningheads]{llncs}
\usepackage[T1]{fontenc}
\usepackage{graphicx}

\usepackage{amsmath,amsfonts,bm}

\def\eqref#1{equation~\ref{#1}}

\def\1{\bm{1}}

\DeclareMathAlphabet{\mathsfit}{\encodingdefault}{\sfdefault}{m}{sl}
\SetMathAlphabet{\mathsfit}{bold}{\encodingdefault}{\sfdefault}{bx}{n}

\usepackage{algorithm}
\usepackage{algorithmic}
\usepackage{subfigure}
\usepackage{hyperref}
\usepackage{wrapfig}

\usepackage[misc]{ifsym}

\usepackage{color}

\begin{document}
\title{Bootstrap State Representation using\\ Style Transfer for Better Generalization \\ in Deep Reinforcement Learning}
\titlerunning{Bootstrap State Representation using Style Transfer}
%
\author{Md Masudur Rahman{\Letter}\inst{1} 
\and
Yexiang Xue\inst{1} }
\authorrunning{M. Rahman and Y. Xue}
%
\institute{Department of Computer Science, Purdue University, West Lafayette IN, 47907, USA \\
\email{\{rahman64,yexiang\}@purdue.edu}\\
}

\tocauthor
\toctitle

\maketitle              
%

\begin{abstract}
Deep Reinforcement Learning (RL) agents often overfit the training environment, leading to poor generalization performance. In this paper, we propose Thinker, a bootstrapping method to remove adversarial effects of confounding features from the observation in an unsupervised way, and thus, it improves RL agents' generalization. Thinker first clusters experience trajectories into several clusters. These trajectories are then bootstrapped by applying a style transfer generator, which translates the trajectories from one cluster's style to another while maintaining the content of the observations. The bootstrapped trajectories are then used for policy learning. Thinker has wide applicability among many RL settings. Experimental results reveal that Thinker leads to better generalization capability in the Procgen benchmark environments compared to base algorithms and several data augmentation techniques. 

\end{abstract}



\section{Introduction}

Deep reinforcement learning has achieved tremendous success. 
However, deep neural networks often overfit to confounding features in the training data due to their high flexibility, leading to poor generalization \cite{hardt2016train,zhang2018dissection,cobbe2019quantifying,cobbe2019procgen}. 
These confounding features (e.g., background color) are usually not connected to the reward; thus, an optimal agent should avoid focusing on them during the policy learning. 
Even worse, confounding features lead to incorrect state representations, which prevents  deep RL agents from performing well even in slightly different environments. 

Many approaches have been proposed to address this challenges including data augmentation approaches such as random cropping, adding jitter in image-based observation \cite{cobbe2019quantifying,laskin2020reinforcement,raileanu2020automatic,kostrikov2020image,laskin2020curl},random noise injection \cite{igl2019generalization}, network randomization \cite{osband2018randomized,burda2018exploration,Lee2020Network}, and regularization \cite{cobbe2019quantifying,kostrikov2020image,igl2019generalization,wang2020improving} have shown to improve generalization.
The common theme of these approaches is to increase diversity in the training data so as the learned policy would better generalize. 
However, this perturbation is primarily done in isolation of the task semantic, which might change an essential aspect of the observation, resulting in sub-optimal policy learning.

\begin{figure}[!ht]
    \centering
    \includegraphics[width=0.90\columnwidth]{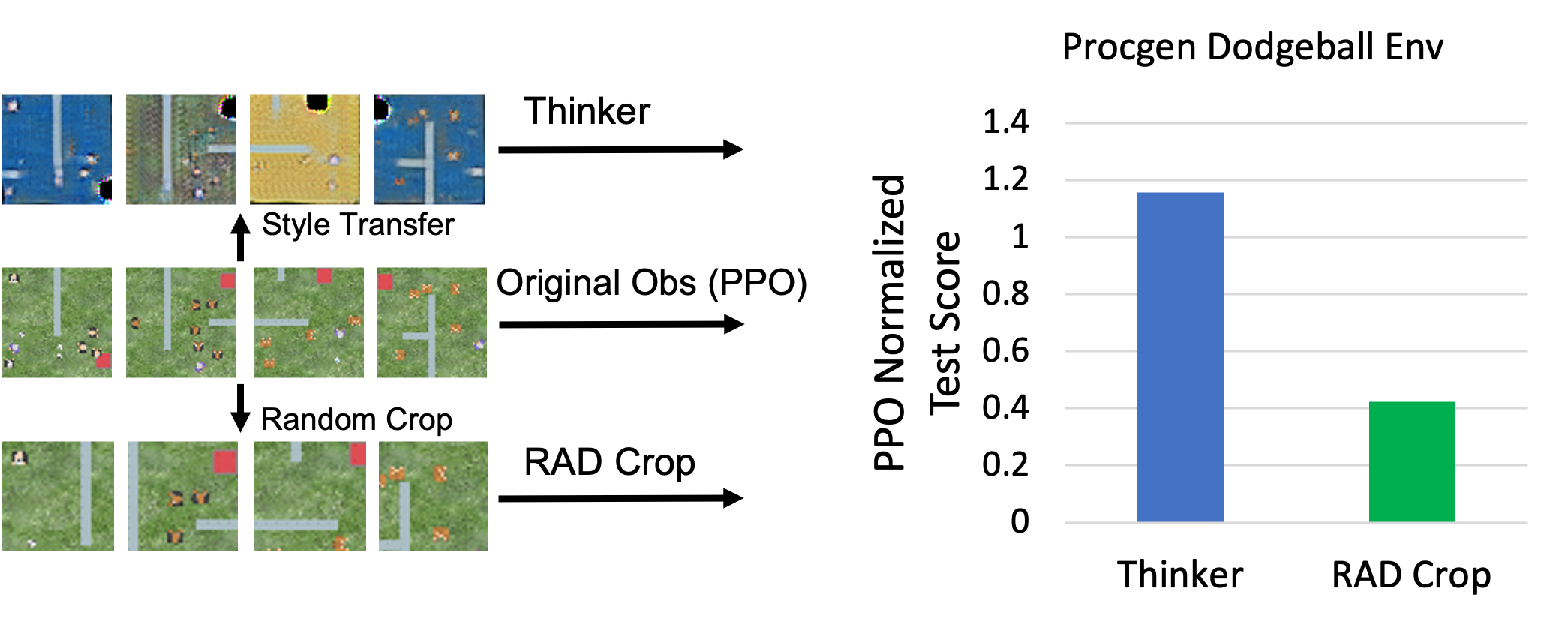}
     \caption{\small Comparison between style transfer-based Thinker and random crop data augmentation-based RAD \cite{laskin2020reinforcement} agents on Procgen Dodgeball. [\textbf{Left}] We see that The random crop removes many essential aspects of the observation while the style transfer retains most game semantics and changes mainly the background and texture of objects. [\textbf{Right}] In generalization, the Thinker agent achieves better performance compared to PPO. In contrast, the RAD Crop agent significantly worsens the base PPO's performance.}
     \label{fig:dodgeball_thinker_overview}
    \centering
\end{figure}

Moreover, the random perturbation in various manipulations of observations such as cropping, blocking, or combining two random images from different environment levels might result in unrealistic observations that the agent will less likely observe during testing. Thus these techniques might work poorly in the setup where agents depend on realistic observation for policy learning. 

For example, consider a RL maze environment where the agent takes the whole maze board image as input observation to learn a policy where the background color of maze varies in each episode. Thus, applying random cropping might hide essential part of the observation which eventually results in poor performance. Our proposed method tackle this issue by changing style of the observation (e.g., background color) while maintaining the maze board's semantic which eventually help RL agent to learn a better policy.
It is also desirable to train the agent with realistic observations, which helps it understand the environments' semantics. Otherwise, the agent might learn unexpected and unsafe behaviors while entirely focusing on maximizing rewards even by exploiting flaws in environments such as imperfect reward design.

In this paper, we propose Thinker, a novel bootstrapping approach to remove the adversarial effects of confounding features from observations and boost the deep RL agent performance.
Thinker automatically creates new training data via changing the visuals of the given observation. 
An RL agent then learns from various observation style instead of a single styled the original training data. Intuitively, this approach help the agent not to focus much on the style which assumed to be confounder and can change in future unseen environments.
Compared to previous approaches, our proposed method focuses on transforming the visual style of observations realistically while keeping the semantics same. 
Thus, the transferred trajectories  corresponds to those that possibly appear in testing environments, hence assisting the agent in adapting to the unseen scenarios. 
Design of our method is motivated by the \textit{counterfactual thinking} nature of human -  \textit{``what if the background color of the image observation was Red instead of Blue?"}; thus the name is Thinker. 
This imagination-based thinking often beneficial for decision-making on similar scenarios in the future events \cite{roese1994functional,epstude2008functional}.

Our method uses a similar mechanism to disentangled confounding features.
Figure \ref{fig:dodgeball_thinker_overview} shows an overview of the Thinker module. 
It maintains a set of distributions (cluster of sample observations) of experience data, which can be learned using a clustering algorithm. 

Our proposed approach consists of a style transfer-based observation translation method that considers content of the observation. Trajectory data from the agent's replay buffer is clustered into different categories, and then observation is translated from one cluster style to another cluster's style. Here the style is determined by the commonality of observation features in a cluster. Thus this style translation is targeted toward non-generalizable features. The agent should be robust toward changes of such features. Moreover, the translated trajectories correspond to those that possibly appear in testing environments, assisting the agent in adapting to unseen scenarios. 

Thinker learns generators between each pair of clusters using adversarial loss \cite{goodfellow2014generative} and cycle consistency loss \cite{zhu2017unpaired,choi2018stargan}. The generator can translate observations from one cluster to another; that means changing style to another cluster while maintaining the semantic of the observation in the underlying task.
After training, all generators are available to the RL agent to use during its policy learning process.

During policy training, the agent can query the Thinker module with new observations and get back the translated observations. The agent can then use the translated observation for policy training. Here, the Thinker module bootstraps the observation data and tries to learn better state representation, which is invariant to the policy network's unseen environment.
Intuitively, the observation translation process is similar to asking the counterfactual question; what if the new observation is coming from a different source (visually different distribution)?

Note that, Thinker works entirely in an unsupervised way and does not require any \textit{additional} environment interactions. Thus the agent can learn policy without collecting more data in the environment, potentially improving sample efficiency and generalization in unseen environments. 

We evaluated the effectiveness of Thinker module on Procgen \cite{cobbe2019procgen} benchmark environments. We evaluated the usefulness of Thinker on the standard on-policy RL algorithm, Proximal Policy Optimization (PPO) \cite{schulman2017proximal}.
We observe that Thinker often can successfully transfers style from one cluster to another, generating semantically equivalent observation data. 
Moreover, our agent performs better in generalization to unseen test environments than PPO.
We further evaluate our method with two popularly used data augmentation approaches: random cropping and random cutout \cite{laskin2020reinforcement}. We demonstrate that these data augmentation method sometimes worsen the base PPO algorithm while our proposed approach improve the performance in both sample efficiency (Train Reward) and generalization (Test Reward).

In summary, our contributions are listed as follows:
\begin{itemize}
    \item We introduce Thinker, a bootstrapping method to remove adverse effects of confounding features from the observation in an unsupervised way.
    \item Thinker can be used with existing deep RL algorithms where experience trajectory is used for policy training. We provide an algorithm to leverage Thinker in different RL settings.
    \item We evaluate Thinker on Procgen environments where it often successfully translates the visual features of observations while keeping the game semantic intact. Overall, our Thinker agent performs better in sample efficiency and generalization than the base PPO \cite{schulman2017proximal} algorithm and and two data augmentation-based approaches: random crop and random cutout \cite{laskin2020reinforcement}. 
\end{itemize}
The source code of our Thinker module is available at \url{https://github.com/masud99r/thinker}.

\section{Background}
\noindent\textbf{Markov Decision Process (MDP)}. 
An MDP can be denoted as $\mathcal{M} =(\mathcal{S}, \mathcal{A}, \mathcal{P}, r)$ where $\mathcal{S}$ is a set states, $\mathcal{A}$ is a set of possible actions. At every timestep $t$, from an state $s_t \in \mathcal{S}$, the agent takes an action $a_t \in \mathcal{A} $ and the environment proceed to next state. The agent then receives a reward $r_t$ as the environment moves to a new state $s_{t+1} \in \mathcal{S}$ based on the transition probability $P(s_{t+1}|s_t, a_t)$. 

\noindent\textbf{Reinforcement Learning}.
In reinforcement learning, the agent interacts with the environment in discrete timesteps that can be defined as an MDP, denoted by $M =(\mathcal{S}, \mathcal{A}, \mathcal{P}, r)$, $\mathcal{P}$ is the transition probability between states after agent takes action, and $r$ is the immediate reward the agent gets. 
In practice, the state ($\mathcal{S}$) is unobserved, and the agent gets to see only a glimpse of the underlying system state in the form of observation ($\mathcal{O}$). 
The agent's target is to learn a policy ($\pi$), which is a mapping from state to action, by maximizing collected rewards. In addition, to master skills in an environment, the agent needs to extract useful information from the observation, which helps take optimal actions.
In deep reinforcement learning (RL), the neural network architecture is often used to represent the policy (value function, Q-function). In this paper, we use such a deep RL setup in image-based observation space.

\noindent\textbf{RL Agent Evaluation}.
Traditionally, RL agent trained in an environment where it is evaluated how quickly it learns the policy. However, the evaluation is often done on the same environment setup. While this evaluation approach can measure policy learning efficiency, it critically misses whether the agent actually learned the necessary skill or just memorized some aspect of the environment to get the maximum reward in training.
In this setup, the agent can often overfit to the scoreboard or timer in a game which can lead to the best reward; however, the agent can completely ignore other parts of the environment \cite{zhang2018study,Song2020Observational}. The agent can even memorize the training environment to achieve the best cumulative reward \cite{zhang2018study}. In contrast, in this paper, we use a zero-shot generalization \cite{Song2020Observational} setup where the agent is trained and tested on different environment instances. Furthermore, the agent's performance is evaluated on unseen environment instances; thus, the agent must master skills during training to perform better in generalization.

\noindent\textbf{Generalization Issue in Deep RL}.
The agent's goal is to use necessary information from the observation and learn behavior that maximizes reward. However, due to the lack of variability in observations, the agent might focus on spurious features. 
This problem becomes commonplace in RL training, especially if the observation space is large, such as the RGB image. In such cases, the agent might memorize the trajectory without actually learning the underlying task. This issue might be undetected if the agent trains and evaluates in the same environment. The agent trained in such a task (environment) might overfit to the trained environment and fail to generalize in the same task but with a slightly different environment.
For example, background color might be irrelevant for a game, and the game might have different backgrounds at different episodes, but the game logic will remain the same. These unimportant features are the confounder that might mislead agents during training. The issue might be severe in deep reinforcement learning as the agent policy is often represented using high-capacity neural networks. If the agent focuses on these confounder features, it might overfit and fails to generalize.

\noindent\textbf{Style Transfer with Generative Adversarial Network}.
The task of style transfer is to change particular features of a given image to another, where generative adversarial network (GAN) has achieved enormous success \cite{kim2017learning,isola2017image,zhu2017unpaired,choi2018stargan}. 
This setup often consists of images from two domains where models learn to style translate images from one domain to another.  The shared features then define the style among images in a domain. A pairing between two domains images is necessary to make many translation methods work. However, such information is not available in the reinforcement learning setup. Nevertheless, an unpaired image-to-image translation method can be used, which does not require a one-to-one mapping of annotated images from two domains. In this paper, we leverage StarGAN \cite{choi2018stargan} that efficiently learns mappings among various domains using a single generator and discriminator. In the RL setup, we apply a clustering approach which first separates the trajectory data into clusters. Then we train the StarGAN on these clusters that learn to style translate images among those clusters.

\section{Bootstrap Observations with Thinker}

\begin{figure}[!ht]
    \centering
    \includegraphics[width=0.95\columnwidth]{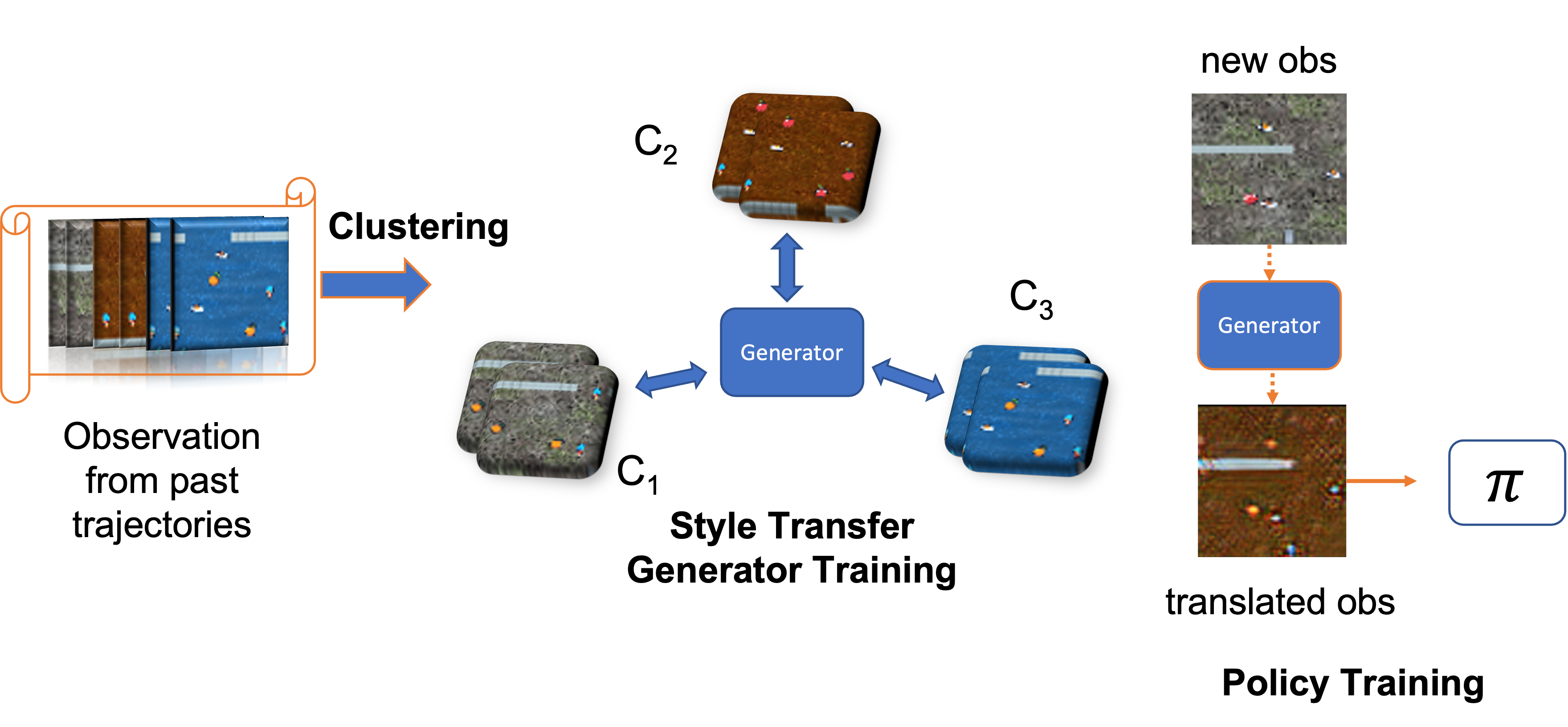}
     \caption{\small  Overview of style Thinker module. The experience trajectory observation data for the task environment are separated into different classes based on visual features using Gaussian Mixture Model (GMM) clustering algorithm. The task of the generator is then to translate the image from one class image to another classes images. In this case, the ``style" is defined as the commonality among images in a single class. Given a new observation, it first infers into its (source) cluster using GMM, and then the generator translated it to (target) another cluster style. The target cluster is taken randomly from the rest of the cluster. The translated observations are used to train the policy.}
     \label{fig:thinker_overview}
    \centering
\end{figure}

Our proposed method, Thinker, focuses on removing the adverse confounding features, which helps the deep RL agent to learn invariant representations from the observations, which eventually help to learn generalizable policy. Figure \ref{fig:thinker_overview} shows an overview of our method.
Thinker maintains a set of distributions achieved by clustering observation data that come from the experience trajectory. 
We implemented our method on a high-dimensional RGB image observation space. 

\noindent\textbf{Clustering Trajectories}.
The trajectory data is first clustered into several ($n$) clusters. Though any clustering algorithms can be leveraged for this clustering process, in this paper, we describe a particular implementation of our method, where we use the Gaussian Mixture Model (GMM) for clustering, and ResNet \cite{he2016deep} for feature extraction and dimension reduction. 
Furthermore, this clustering process focuses entirely on the visual aspect of the observation without necessarily concentrating on the corresponding reward structure. Therefore, images would be clustered based on these visual characteristics.
In the next step, the observation dataset is clustered using the GMM algorithm. Images in these clusters' are then used to carry out style transfer training.

\noindent\textbf{Generator Training}.
We train a single generator $G$ to translate image from one cluster to another. We build on the generator on previous works \cite{choi2018stargan,zhu2017unpaired} which is a unified framework for a multi-domain image to image translation.
Given an input image $x$ from a source cluster the output translated image $x'$ conditioned on the target cluster number $c$, that is $ x' \leftarrow G(x,c)$, where $c$ is a randomly chosen  cluster number. A discriminator $D: x \rightarrow \{D_{src}(x), D_{cls}(x)\}$ is used to distinguish real image and fake image generated by $G$. Here $D_{src}$ distinguish between fake and real images of the source, and $D_{cls}$ determines the cluster number of the given input image $x$.
Generator G tries to fool discriminator $D$ in an adversarial setting by generating a realistic image represented by the true image distribution. 

The adversarial loss is calculated as the Wasserstein GAN \cite{arjovsky2017wasserstein} objective with gradient penalty  \cite{gulrajani2017improved} which stabilize the training compared to regular GAN objective \cite{goodfellow2014generative}. This loss is defined as 
\begin{equation} \label{eq:loss_disciminator}
    \mathcal{L}_{adv} = \mathbb{E}_x[D_{src}] - \mathbb{E}_{x,c}[D_{src}(G(x,c))] - \lambda_{gp} \mathbb{E}_{\hat{x}}[(||\nabla_{\hat{x}}D_{src}(\hat{x})|| - 1)^2],
\end{equation}
where $\hat{x}$ is sampled uniformly along a straight line between a pair of real and generated fake images and $\lambda_{gp}$ is a hyperparameter.
The cluster classification loss is defined for real and fake images.
The classification loss of real image is defined as 
\begin{equation} \label{eq:loss_class_real}
    \mathcal{L}^r_{cls} = \mathbb{E}_{x, c'}[-\log D_{cls}(c'|x)]
    ,
\end{equation}
where $D_{cls}(c'|x)$ is the probability distribution over all cluster labels.
Similarly, the classification loss of fake generated image is defined as 
\begin{equation} \label{eq:loss_class_real}
    \mathcal{L}^f_{cls} = \mathbb{E}_{x, c}[-\log D_{cls}(c|G(x, c))]
    ,
\end{equation}

The full discriminator loss is
\begin{equation} \label{eq:loss_disciminator}
    \mathcal{L}_D = -\mathcal{L}_{adv} + \lambda_{cls} \mathcal{L}^r_{cls},
\end{equation}
which consists of the adversarial loss $\mathcal{L}_{adv}$, and domain classification loss $\mathcal{L}^r_{cls}$ and $\lambda_{cls}$ is a hyperparameter. The discriminator detects a fake image generated by the generator G from the real image in the given class data.

To preserve image content during translation a reconstruction loss is applied
\begin{equation} \label{eq:loss_generator}
    \mathcal{L}_{rec} = \mathbb{E}_{x, c, c'}[||x-G(G(x,c), c')||_1], 
\end{equation}
where we use the $L1$ norm.

The $\mathcal{L}_{rec}$ is the reconstruction loss which makes sure the generator preserves the content of the input images while changing the domain-related part of the inputs. This cycle consistency loss \cite{choi2018stargan,zhu2017unpaired} $\mathcal{L}_{rec}$ makes sure the translated input can be translated back to the original input, thus only changing the domain related part and not the semantic.
Thus, the generator loss is
\begin{equation} \label{eq:loss_generator}
    \mathcal{L}_G = \mathcal{L}_{adv} + \lambda_{cls} \mathcal{L}^f_{cls} +\lambda_{rec} \mathcal{L}_{rec},
\end{equation}
where $\mathcal{L}_{adv}$ is adversarial loss, and  $\mathcal{L}^f_{cls}$ is the loss of detecting fake image and the $\lambda_{rec}$ is a hyperparameter.

\begin{algorithm}
    \caption{Thinker}
    \label{algo-thinker}
    \begin{algorithmic}
    \STATE Get PPO for policy learning RL agent
    \STATE Collect  observation trajectory $\mathcal{D}$ using initial policy
        \STATE Cluster dataset $\mathcal{D}$ into $n$ clusters using GMM
        \STATE Train Generator $G$ with the $n$ clusters by optimizing equation \ref{eq:loss_disciminator}, and \ref{eq:loss_generator}
        \FOR {each iteration}
    
        \FOR{each environment step} 
             \STATE $a_t \sim \pi_\theta(a_t|x_t)$ 
            \STATE $x_{t+1} \sim P(x_{t+1}|x_t, a_t)$ 
            \STATE $r_t \sim R(x_t, a_t)$ 
            \STATE $\mathcal{B} \xleftarrow{} \mathcal{B} \cup \{(x_t, a_t, r_t, x_{t+1})\} $ 
             \STATE Translate all obs $x \in \mathcal{B}$ to  get $\mathcal{B}'$ using Generator $G$
            
            \STATE Train policy $\pi_\theta$ on $\mathcal{B}'$ with PPO
        \ENDFOR
        \ENDFOR
       
    \end{algorithmic}
    \end{algorithm}

\noindent\textbf{Train Agent with Thinker}. 
During policy training, the agent can query the generator module with a new observation and get back translated observation (Algorithm \ref{algo-thinker}). The agent can then use the translated observation for policy training. 
Intuitively, the observation translation process is similar to asking the counterfactual question; \textit{``what if the new observation is coming from a visually different episode distribution)?"}
The Thinker method can be applied to existing deep RL algorithms where experience data is used to train policy networks. In this paper, we evaluate Thinker with on-policy PPO \cite{schulman2017proximal}.
Intuitively, Thinker maintains a counterfactual-based visual thinking component, which it can invoke at any learning timestep and translate the observation from one distribution to another. Algorithm \ref{algo-thinker} describes detailed steps of training deep RL agents with Thinker.

\section{Experiments}
\subsection{Implementation}
We implemented Thinker using Ray framework:
Tune, and RLlib \cite{liang2018rllib}, which supports simple primitive and unified API to build scalable applications.

\noindent\textbf{Clustering}.
We use Gaussian Mixture Model (GMM) implementation available in Scikit-learn \cite{scikit-learn}. 
We first pass observation through the pre-trained ResNet18 \cite{he2016deep} model which is trained on ImageNet dataset \cite{ILSVRC15}. 
The ResNet18 model \footnote{\url{https://pytorch.org/hub/pytorch_vision_resnet/}} converts the RGB image into a 1000 dimensional vector. We use the layer just before the final softmax layer to get this vector. This dimensionality reduction step drastically reduces the training and inference time of the Gaussian mixture model. 
Given the number of cluster $n$, this model train on $n$ clusters. These $n$ clusters data are stored by the agent, which is later used for the generator training. The inference module takes input an observation, and it returns the cluster-ID to which it belongs, which is used to identify the target cluster for the style translation.
The number of clusters is the hyperparameter, which can depend on the diversity of environment levels. 
However, for our method to be effective, at least two clusters are required. Therefore, unless otherwise mentioned, we reported comparison results using the number of clusters $n=3$ (better performing in ablation).

\noindent\textbf{Learning Generator}.
After clustering, all data is then feed to the generator module which learns a single generator that style translate between any pair of clusters.
The agent can choose various cluster numbers (hyperparameter) during training time. Each time the clustering is trained, the generator must be updated with the new cluster samples. 
In our experiments, we train the generator once and at the beginning of the training.
Note that the collected trajectory data should be representative enough to train a good generator. Thus initially, the agent has to sufficiently explore the environments to have a diverse observation in the buffer. In our experiment, we use an initial policy whose parameters are chosen randomly to allow exploration and enable diverse data collection for the cluster and generator training.

\subsection{Experiment Setup}
  \begin{figure}[!ht]
    \centering
    \includegraphics[width=0.70\linewidth]{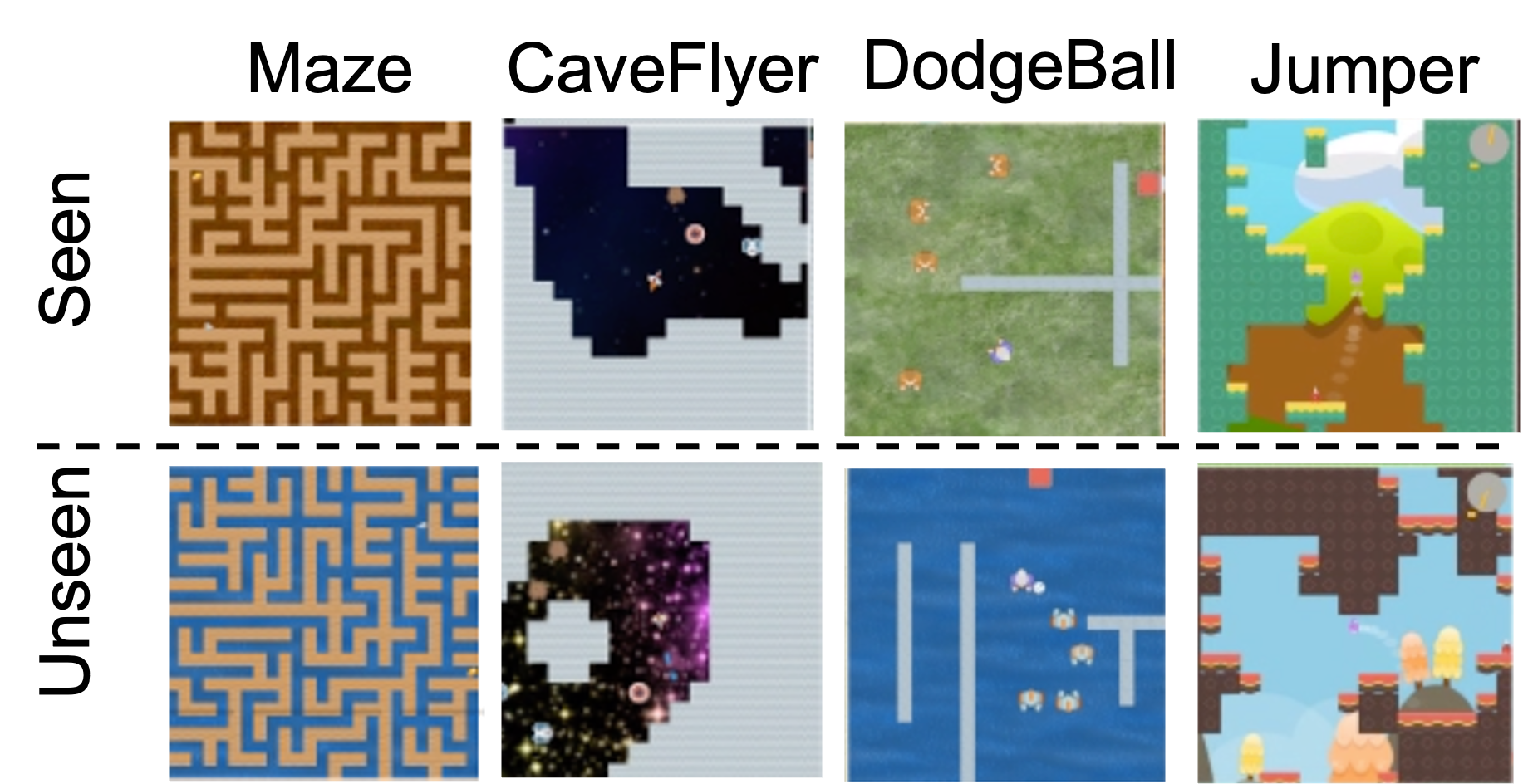} 
    \caption{\small  
     Some snippets of different Procgen environments. The training (seen) levels vary drastically from the testing (unseen) environment. The agent must master the skill without overfitting irrelevant non-generalizable aspects of the environment to perform better in unseen levels.
    }
    \centering
    \label{fig:procgen_env}
    \end{figure}
For the StarGAN training, we use 500 iterations, where in each iteration, the data were sampled from the available clusters dataset. These sampled data were used to train the generators' networks.
For the generator model, we use a ResNet-based CNN architecture with 6 residual blocks. The hyperparameters are set $\lambda_{cls}=1$, $\lambda_{rec}=10$, and $\lambda_{gp}=10$.

\textbf{Environment}. We conducted experiments on four OpenAI Procgen \cite{cobbe2019procgen} environments consisting of diverse procedurally-generated environments with different action sets: Maze, CaveFlyer, Dodgeball, and Jumper. These environments are chosen due to their relatively larger generalization gap \cite{cobbe2019procgen}.
We conduct experiments on these environments to measure how quickly (sample efficiency) a reinforcement learning agent learns generalizable policy. 
Some snippets of different Procgen environments are given in Figure \ref{fig:procgen_env}. 
All environments use a discrete 15 dimensional action space which generates $64 \times 64 \times 3$  RGB image observations. 

\noindent\textbf{Settings}.
As suggested in the Procgen benchmark paper \cite{cobbe2019procgen}, we trained the agents on 200 levels for \textit{easy} difficulty levels and evaluated on the full distribution of levels.
We report evaluation results on the full distribution (i.e., test), including unseen levels, focusing on generalization as well as training learning curve for sample efficiency.
We used the standard Proximal Policy Optimization (PPO) \cite{schulman2017proximal} and data augmentation techniques for our baseline comparison.
PPO learns policy in an on-policy approach by alternating between sampling data through interaction with the environment and optimizing a surrogate objective function, enabling multiple epochs of minibatch updates using stochastic gradient ascent.

On the other hand, RAD is a 
data augmentation technique \cite{laskin2020reinforcement} which shows effective empirical evidence in complex RL benchmarks including some Procgen environments. 
In particular, the \textit{Cutout Color} augmentation technique which has shown better results in many Procgen environments compared in \cite{laskin2020reinforcement} thus we compare with this data augmentation technique.
Additionally, we experimented on random crop augmentation. 
However, this augmentation fails to achieve any reasonable performance in the experimented environments. Thus, we do not report the results for random crop here in our experiments.

We used RLlib \cite{liang2018rllib} to implement all the algorithms. For all the agents (Thinker, PPO, and RAD), to implement the policy network (model), we use a CNN architecture used in IMPALA \cite{espeholt2018impala}, which also found to work better in the Procgen environments \cite{cobbe2019procgen}. 
To account for the agents' performance variability, we run each algorithm with 5 random seeds. 
Policy learning hyperparameter settings (RLlib's default\cite{liang2018rllib}) for Thinker, PPO, and RAD are set the same for a fair comparison. The hyperparameters are given in Table \ref{tab:rllib-hyp}. 

\begin{table}
\caption{Hyperparameters for Experiments - RLlib} 
\label{tab:rllib-hyp} 
\begin{center}
 \scriptsize
\begin{tabular}{|c|c||c|c|}
\hline
 Description & Hyperparameters &  Description & Hyperparameters  \\
\hline
 Discount factor & $0.999$ & The GAE(lambda) & $0.95$  \\
\hline
Learning rate & $5.0e-4$ & Epochs per train batch & $3$ \\
\hline
SGD batch & $2048$ & Training batch size & $16384$\\
\hline
KL divergence &  $0.0$ & Target KL divergence & $0.01$\\
\hline
Coeff. of value loss & $0.5$ & Coeff. of the entropy & $0.01$ \\
\hline
PPO clip parameter & $0.2$ & Clip for the value & $0.2$\\
\hline
Global clip & $0.5$ & PyTorch Framework & $torch$\\
\hline
Settings for Model & IMPALA CNN & Rollout Fragment & $256$ \\
\hline
\end{tabular}
\end{center}
\end{table}

\noindent\textbf{Evaluation Metric}.
It has been observed that a single measure in the form of mean or median can hide the uncertainty implied by different runs \cite{agarwal2021deep}. In this paper, we report the reward distribution of all 5 random seed runs in the form of a boxplot to mitigate the above issue.

\noindent\textbf{Computing details}.
We used the following machine configurations to run our experiments: 20 core-CPU with 256 GB of RAM, CPU Model Name: Intel(R) Xeon(R) Silver 4114 CPU @ 2.20GHz, and an Nvidia A100 GPU. 
In our setup, for each run of a training of 25M timesteps, Thinker took approx. 14 hours (including approx. 2 hours of generator training), RAD-Random Crop took approx. 30 hours, RAD-Cutout Color took approx. 9 hours and PPO took approx. 8 hours.
\subsection{Results}
We now discuss the results of our experiments. We first discuss the generalization results and then sample efficiency. Further, we evaluate how our agents perform in different hyperparameter values of the number of clusters. Finally, we demonstrate samples of the style transfer by our generator.

\noindent\textbf{Generalization on unseen environments}.
\begin{figure}
  \centering
  \begin{minipage}[b]{0.24\textwidth}
    \includegraphics[width=0.99\textwidth]{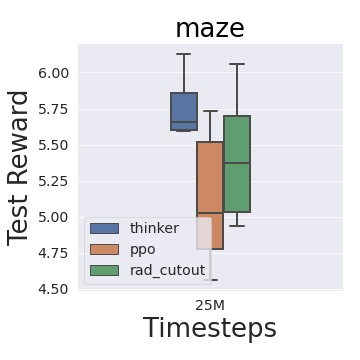}
  \end{minipage}
  \hfill
  \begin{minipage}[b]{0.24\textwidth}
    \includegraphics[width=0.99\textwidth]{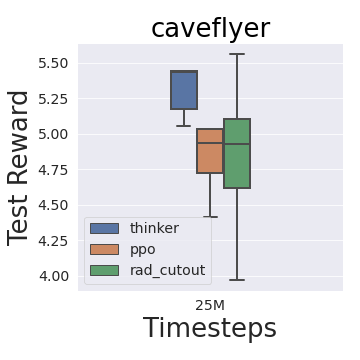}
  \end{minipage}
   \hfill
  \begin{minipage}[b]{0.24\textwidth}
    \includegraphics[width=0.99\textwidth]{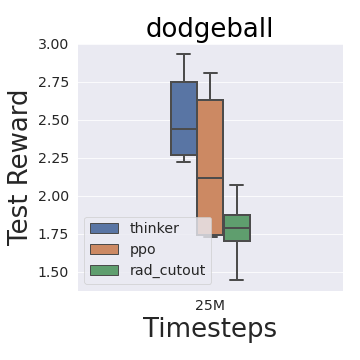}
  \end{minipage}
  \hfill
  \begin{minipage}[b]{0.24\textwidth}
    \includegraphics[width=0.99\textwidth]{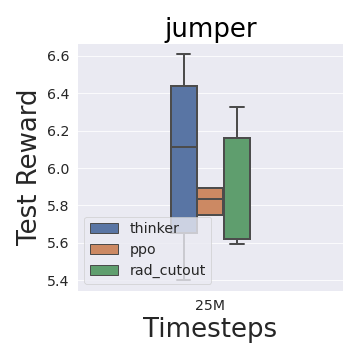}
  \end{minipage}

  \caption{ Generalization (Test) results. Our agent Thinker performs better in all environments than the base PPO algorithm and RAD cutout data augmentation.}
  \label{fig:thinker_procgen_test}
\end{figure}
\begin{figure}
  \centering
  \begin{minipage}[b]{0.24\textwidth}
    \includegraphics[width=0.99\textwidth]{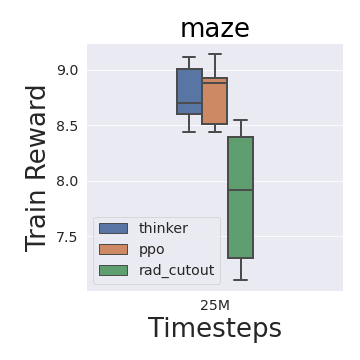}
  \end{minipage}
  \hfill
  \begin{minipage}[b]{0.24\textwidth}
    \includegraphics[width=0.99\textwidth]{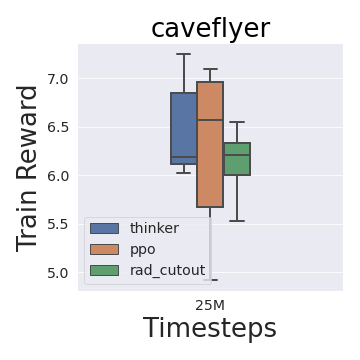}
  \end{minipage}
  \hfill
  \begin{minipage}[b]{0.24\textwidth}
    \includegraphics[width=0.99\textwidth]{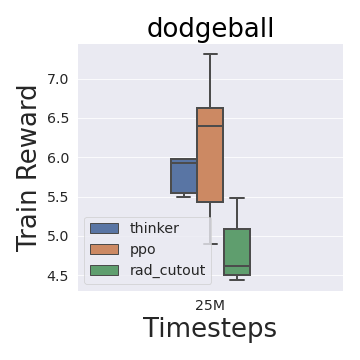}
  \end{minipage}
  \hfill
  \begin{minipage}[b]{0.24\textwidth}
    \includegraphics[width=0.99\textwidth]{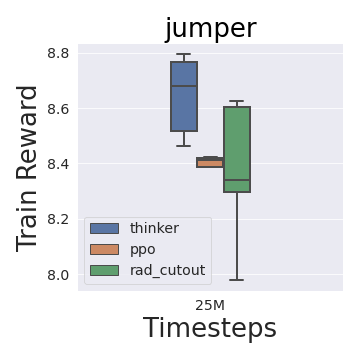}
  \end{minipage}
  \caption{Sample efficiency (Train) results.
  Thinker achieves better sample efficiency in the Jumper environment while performing comparably with the base PPO algorithm in other environments. Note that our agent Thinker still achieves competitive results during training despite being optimized for generalization.
  }
  \label{fig:thinker_procgen_train}
\end{figure}
We show how each agent achieves generalization after training for 25 million timesteps. This scenario is a zero-shot setting, which means we do not train the agent on the test environment levels (unseen to the trained agent). We report the reward in different random seed runs in a boxplot. The generalization results are computed by evaluating the trained agents on test levels (full distribution) for 128 random episode trials.

Figure \ref{fig:thinker_procgen_test} shows the boxplot of the test performance at the end of the training. We observe that our agent Thinker performs better (in the median, 25th, and 75th reward) compared to the base PPO algorithm and RAD cutout data augmentation. On the other hand, the random cutout data augmentation approach sometimes worsens the performance compared to the base PPO. In all cases, Thinker performs better than the data augmentation-based approach. Random Crop performed worst and could not produce any meaningful reward in these environments. Thus, for brevity, we omit them from Figure \ref{fig:thinker_procgen_test}.

These results show the importance of our bootstrapped observations data during policy training, which could help us learn a policy that performs better across unseen levels of environments than baselines.

\noindent\textbf{Sample efficiency during training}.
We further evaluate the sample efficiency of our method during training. We show in Figure \ref{fig:thinker_procgen_train} the final train reward after training the agents for 25 million timesteps.

 \begin{figure}
  \centering
  \begin{minipage}[b]{0.55\textwidth}
    \includegraphics[width=0.90\textwidth]{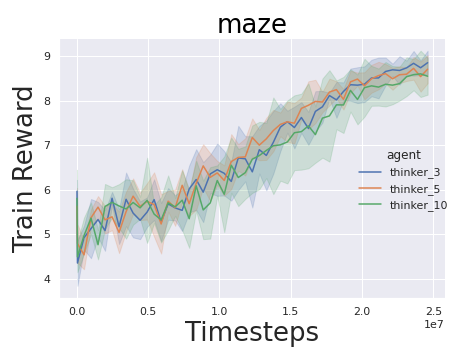}
  \end{minipage}
  \hfill
  \begin{minipage}[b]{0.44\textwidth}
    \includegraphics[width=0.90\textwidth]{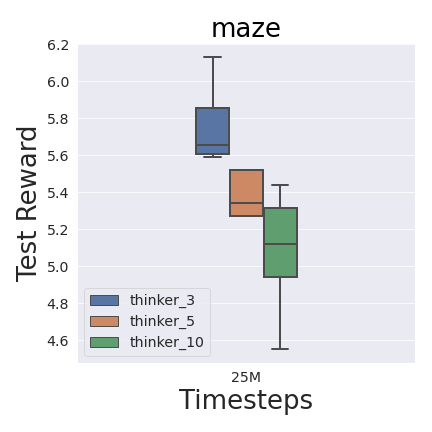}
  \end{minipage}
  \hfill
  \caption{Ablation results. Thinker's performance on different cluster numbers (3, 5, and 10) on the Maze Procgen environment. The results are averaged over 5 seeds. [\textbf{Left}] Thinker's learning curve during training. [\textbf{Right}] Thinker's generalization performance in boxplot on unseen levels after the training.}
  \label{fig:thinker_procgen_ablation}
\end{figure}
Thinker achieves better sample efficiency in the Jumper environment and performs comparably with the base PPO algorithm in other environments. However, the random cutout data augmentation mostly fails to improve (and sometimes worsens) the performance over the base PPO algorithm. Note that the ultimate goal of our agent Thinker is to perform better in test time. Despite that objective, it still achieves competitive results during training.

On the other hand, our agent Thinker performs better than the data augmentation-based approach in all the environments. We omit the random crop data augmentation result for brevity due to its poor performance.
\ref{fig:thinker_procgen_ablation}. 

\noindent\textbf{Ablation Study}.
The ablation results for different cluster numbers are shown in Figure \ref{fig:thinker_procgen_ablation}.
We observe that the number of clusters has some effect on policy learning. The generalization (Test Reward) performance is dropping with the increase in clusters.
When the number of clusters is large, that is 10; the generator might overfit each cluster's features and translate the essential semantic part of the observation, thus resulting in lower performance. However, the cluster number does not affect the train results (Train Reward). We see the best results at cluster number 3 in the Maze Procgen environment.

\begin{figure}[]
\centering
\includegraphics[width=0.90\linewidth]{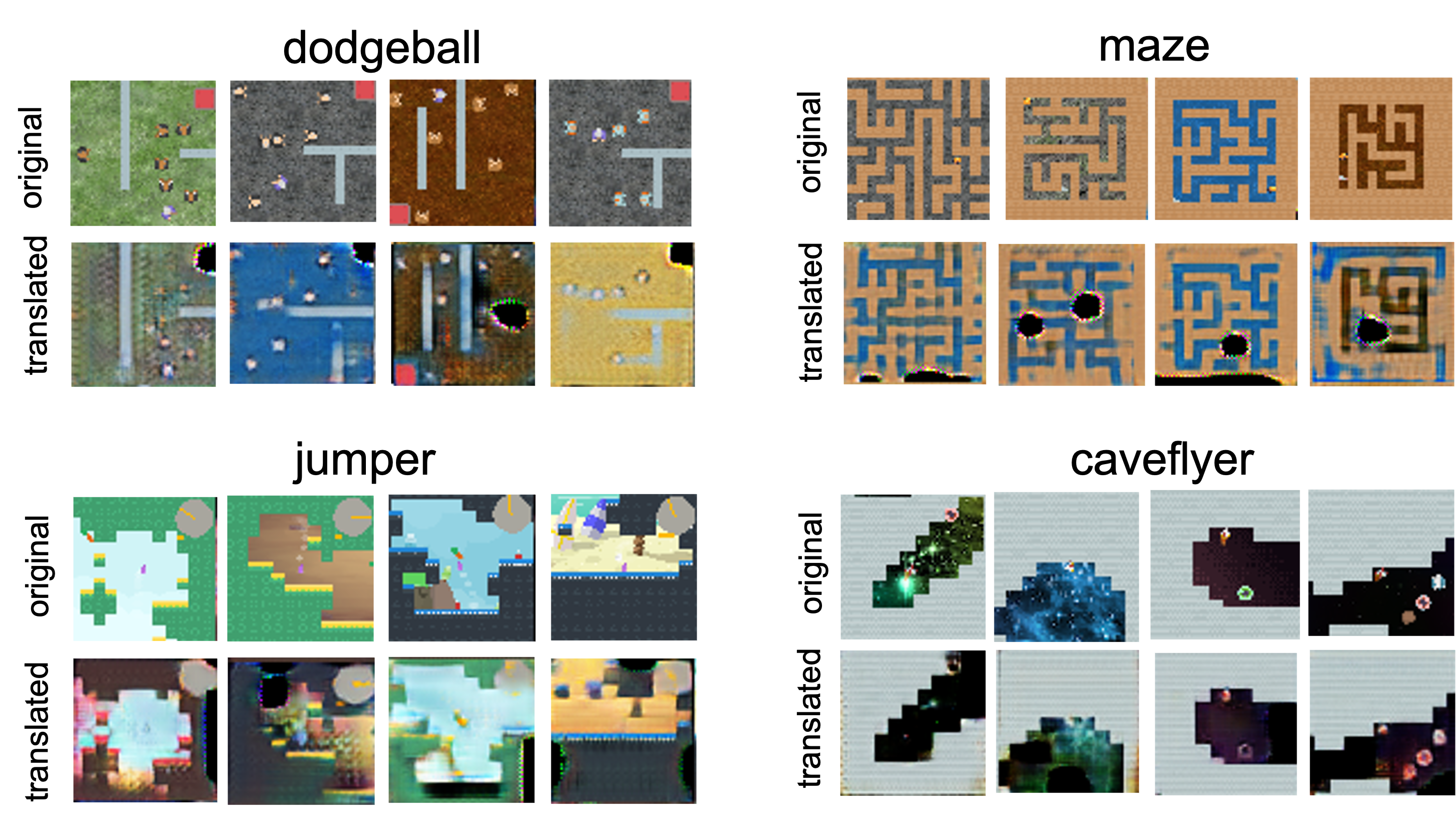} 
\caption{\small 
Sample style translations by our trained generator on some Procgen environments. The top images are the original observations for each environment, and the corresponding bottom images are the translated images. We see that the contents of the translated images mostly remain similar to the original images while the style varies.
}
\centering
\label{fig:translation_sample}
\end{figure}

\noindent\textbf{Style Transfer Sample}. Figure \ref{fig:translation_sample} shows some sample style translations by our trained generator. 
Overall, the generator performs style transfer while mostly maintaining the game semantics. For example, in the Dodgeball environment, in the second column, we see that the background color of the observation is gray, while in the translated observation, it is mostly blue. Additionally, the game objects (e.g., small dots, horizontal and vertical bar) remain in place. These objects are the essential part where the agent needs to focus while solving the task.

\section{Related Work} 
Regularization has been used to improve RL generalization \cite{farebrother2018generalization,cobbe2019quantifying,kostrikov2020image,igl2019generalization,wang2020improving}. On the other hand, data augmentation has been shown promising results in generalization and  in high-dimensional observation space \cite{cobbe2019quantifying,laskin2020reinforcement,kostrikov2020image,raileanu2020automatic,kostrikov2020image}. 
Network randomization \cite{osband2018randomized,burda2018exploration,Lee2020Network} and random noise injection \cite{igl2019generalization}, leveraging inherent sequential structure in RL \cite{agarwal2021pse} have been explored to improve RL robustness and generalization. In these cases, the idea is to learn invariant features and disentangled representation \cite{higgins2017darla} robust to visual changes in the testing environment. 
In our work, we explicitly tackle this problem by generating semantically similar but visually different observation samples, which ideally cancel out unimportant features in the environment and thus learn invariant state representations. Our method focuses on performing realistic visual style transfer of observations while keeping the semantic same. Thus, the target observation corresponds to possible testing environments, aiming to prepare the agent for the unseen scenarios. 
A closely related paper of our style transfer approach is \cite{gamrian2019transfer}. 
They require access to the annotated agent's trajectory data in both source and target domains for the GAN training.
In our case, we do not need the information of levels and their sample beforehand; instead, we automatically cluster the trajectory data based on observation's visual features. 
Thus, the style transfer happens between learned clusters.
Additionally, our approach uses visual-based clustering; thus, one cluster may have data from multiple levels, potentially preventing GAN from overfitting \cite{gamrian2019transfer} to any particular environment levels.

\section{Discussion}
In conclusion, we proposed a novel bootstrapping method to remove the adverse effects of confounding features from the observation in an unsupervised way. 
Our method first clusters experience trajectories into several clusters; then, it learns StarGAN-based generators. These generators translate the trajectories from one cluster's style to another, which are used for policy training. Our method can be used with existing deep RL algorithms where experience trajectory is used for policy training. Evaluating on visually enriched environments, we demonstrated that our method improves the performance of the existing RL algorithm while achieving better generalization capacity and sample efficiency.

The impacts of Thinker on policy learning depends on the quality of the bootstrapped data generations. Thus our method is better suited for the cases where different levels of an environment vary visually (e.g., changing background color, object colors, texture). In the scenarios where different levels of an environment vary due to mostly its semantic logic differences (e.g., the structural difference in a maze), our method might face challenges.
Lack of visual diversity in the clustering might lead the generator to overfit, impacting the its translation performance across these clusters. A possible alternative is to cluster observation data using other features that vary between clusters in addition to visual aspects. A large number of the cluster might place less diverse observation in individual cluster focusing on low-level objects' details, which might cause the generator to overfit. We suggest to reduce the number of clusters in such scenarios.
During policy learning, the agent requires some time to train and infer the Thinker module. However, this additional time is negligible compared to deep RL agents' typical stretched running time. 
Additionally, as we are training a single generator for all the cluster pairs, we find the overhead of Thinker training time reasonable in the context of deep RL agent training.

\subsubsection{Acknowledgements} 
This research was supported by
NSF grants IIS-1850243, CCF-1918327.

%
%
%
\bibliographystyle{splncs04}
\bibliography{main}

\begin{thebibliography}{10}
\providecommand{\url}[1]{\texttt{#1}}
\providecommand{\urlprefix}{URL }
\providecommand{\doi}[1]{https://doi.org/#1}

\bibitem{agarwal2021pse}
Agarwal, R., Machado, M.C., Castro, P.S., Bellemare, M.G.: Contrastive
  behavioral similarity embeddings for generalization in reinforcement
  learning. In: International Conference on Learning Representations (2021)

\bibitem{agarwal2021deep}
Agarwal, R., Schwarzer, M., Castro, P.S., Courville, A., Bellemare, M.G.: Deep
  reinforcement learning at the edge of the statistical precipice. Advances in
  Neural Information Processing Systems  (2021)

\bibitem{arjovsky2017wasserstein}
Arjovsky, M., Chintala, S., Bottou, L.: Wasserstein generative adversarial
  networks. In: International conference on machine learning. pp. 214--223.
  PMLR (2017)

\bibitem{burda2018exploration}
Burda, Y., Edwards, H., Storkey, A., Klimov, O.: Exploration by random network
  distillation. arXiv preprint arXiv:1810.12894  (2018)

\bibitem{choi2018stargan}
Choi, Y., Choi, M., Kim, M., Ha, J.W., Kim, S., Choo, J.: Stargan: Unified
  generative adversarial networks for multi-domain image-to-image translation.
  In: Proceedings of the IEEE conference on computer vision and pattern
  recognition. pp. 8789--8797 (2018)

\bibitem{cobbe2019procgen}
Cobbe, K., Hesse, C., Hilton, J., Schulman, J.: Leveraging procedural
  generation to benchmark reinforcement learning. arXiv preprint
  arXiv:1912.01588  (2019)

\bibitem{cobbe2019quantifying}
Cobbe, K., Klimov, O., Hesse, C., Kim, T., Schulman, J.: Quantifying
  generalization in reinforcement learning. In: International Conference on
  Machine Learning. pp. 1282--1289. PMLR (2019)

\bibitem{epstude2008functional}
Epstude, K., Roese, N.J.: The functional theory of counterfactual thinking.
  Personality and social psychology review  \textbf{12}(2),  168--192 (2008)

\bibitem{espeholt2018impala}
Espeholt, L., Soyer, H., Munos, R., Simonyan, K., Mnih, V., Ward, T., Doron,
  Y., Firoiu, V., Harley, T., Dunning, I., et~al.: Impala: Scalable distributed
  deep-rl with importance weighted actor-learner architectures. arXiv preprint
  arXiv:1802.01561  (2018)

\bibitem{farebrother2018generalization}
Farebrother, J., Machado, M.C., Bowling, M.: Generalization and regularization
  in dqn. arXiv preprint arXiv:1810.00123  (2018)

\bibitem{gamrian2019transfer}
Gamrian, S., Goldberg, Y.: Transfer learning for related reinforcement learning
  tasks via image-to-image translation. In: International Conference on Machine
  Learning. pp. 2063--2072. PMLR (2019)

\bibitem{goodfellow2014generative}
Goodfellow, I., Pouget-Abadie, J., Mirza, M., Xu, B., Warde-Farley, D., Ozair,
  S., Courville, A., Bengio, Y.: Generative adversarial nets. Advances in
  neural information processing systems  \textbf{27},  2672--2680 (2014)

\bibitem{gulrajani2017improved}
Gulrajani, I., Ahmed, F., Arjovsky, M., Dumoulin, V., Courville, A.: Improved
  training of wasserstein gans. arXiv preprint arXiv:1704.00028  (2017)

\bibitem{hardt2016train}
Hardt, M., Recht, B., Singer, Y.: Train faster, generalize better: Stability of
  stochastic gradient descent. In: International Conference on Machine
  Learning. pp. 1225--1234. PMLR (2016)

\bibitem{he2016deep}
He, K., Zhang, X., Ren, S., Sun, J.: Deep residual learning for image
  recognition. In: Proceedings of the IEEE conference on computer vision and
  pattern recognition. pp. 770--778 (2016)

\bibitem{higgins2017darla}
Higgins, I., Pal, A., Rusu, A.A., Matthey, L., Burgess, C.P., Pritzel, A.,
  Botvinick, M., Blundell, C., Lerchner, A.: Darla: Improving zero-shot
  transfer in reinforcement learning. arXiv preprint arXiv:1707.08475  (2017)

\bibitem{igl2019generalization}
Igl, M., Ciosek, K., Li, Y., Tschiatschek, S., Zhang, C., Devlin, S., Hofmann,
  K.: Generalization in reinforcement learning with selective noise injection
  and information bottleneck. In: Advances in neural information processing
  systems. pp. 13978--13990 (2019)

\bibitem{isola2017image}
Isola, P., Zhu, J.Y., Zhou, T., Efros, A.A.: Image-to-image translation with
  conditional adversarial networks. In: Proceedings of the IEEE conference on
  computer vision and pattern recognition. pp. 1125--1134 (2017)

\bibitem{kim2017learning}
Kim, T., Cha, M., Kim, H., Lee, J.K., Kim, J.: Learning to discover
  cross-domain relations with generative adversarial networks. In:
  International Conference on Machine Learning. pp. 1857--1865. PMLR (2017)

\bibitem{kostrikov2020image}
Kostrikov, I., Yarats, D., Fergus, R.: Image augmentation is all you need:
  Regularizing deep reinforcement learning from pixels. arXiv preprint
  arXiv:2004.13649  (2020)

\bibitem{laskin2020reinforcement}
Laskin, M., Lee, K., Stooke, A., Pinto, L., Abbeel, P., Srinivas, A.:
  Reinforcement learning with augmented data. In: Advances in neural
  information processing systems (2020)

\bibitem{laskin2020curl}
Laskin, M., Srinivas, A., Abbeel, P.: Curl: Contrastive unsupervised
  representations for reinforcement learning. In: International Conference on
  Machine Learning. pp. 5639--5650. PMLR (2020)

\bibitem{Lee2020Network}
Lee, K., Lee, K., Shin, J., Lee, H.: Network randomization: A simple technique
  for generalization in deep reinforcement learning. In: International
  Conference on Learning Representations (2020)

\bibitem{liang2018rllib}
Liang, E., Liaw, R., Nishihara, R., Moritz, P., Fox, R., Goldberg, K.,
  Gonzalez, J., Jordan, M., Stoica, I.: Rllib: Abstractions for distributed
  reinforcement learning. In: International Conference on Machine Learning. pp.
  3053--3062. PMLR (2018)

\bibitem{osband2018randomized}
Osband, I., Aslanides, J., Cassirer, A.: Randomized prior functions for deep
  reinforcement learning. In: Advances in Neural Information Processing
  Systems. pp. 8617--8629 (2018)

\bibitem{scikit-learn}
Pedregosa, F., Varoquaux, G., Gramfort, A., Michel, V., Thirion, B., Grisel,
  O., Blondel, M., Prettenhofer, P., Weiss, R., Dubourg, V., Vanderplas, J.,
  Passos, A., Cournapeau, D., Brucher, M., Perrot, M., Duchesnay, E.:
  Scikit-learn: Machine learning in {P}ython. Journal of Machine Learning
  Research  \textbf{12},  2825--2830 (2011)

\bibitem{raileanu2020automatic}
Raileanu, R., Goldstein, M., Yarats, D., Kostrikov, I., Fergus, R.: Automatic
  data augmentation for generalization in deep reinforcement learning. arXiv
  preprint arXiv:2006.12862  (2020)

\bibitem{roese1994functional}
Roese, N.J.: The functional basis of counterfactual thinking. Journal of
  personality and Social Psychology  \textbf{66}(5), ~805 (1994)

\bibitem{ILSVRC15}
Russakovsky, O., Deng, J., Su, H., Krause, J., Satheesh, S., Ma, S., Huang, Z.,
  Karpathy, A., Khosla, A., Bernstein, M., Berg, A.C., Fei-Fei, L.: {ImageNet
  Large Scale Visual Recognition Challenge}. International Journal of Computer
  Vision (IJCV)  \textbf{115}(3),  211--252 (2015)

\bibitem{schulman2017proximal}
Schulman, J., Wolski, F., Dhariwal, P., Radford, A., Klimov, O.: Proximal
  policy optimization algorithms. arXiv preprint arXiv:1707.06347  (2017)

\bibitem{Song2020Observational}
Song, X., Jiang, Y., Tu, S., Du, Y., Neyshabur, B.: Observational overfitting
  in reinforcement learning. In: International Conference on Learning
  Representations (2020)

\bibitem{wang2020improving}
Wang, K., Kang, B., Shao, J., Feng, J.: Improving generalization in
  reinforcement learning with mixture regularization. arXiv preprint
  arXiv:2010.10814  (2020)

\bibitem{zhang2018dissection}
Zhang, A., Ballas, N., Pineau, J.: A dissection of overfitting and
  generalization in continuous reinforcement learning. arXiv preprint
  arXiv:1806.07937  (2018)

\bibitem{zhang2018study}
Zhang, C., Vinyals, O., Munos, R., Bengio, S.: A study on overfitting in deep
  reinforcement learning. arXiv preprint arXiv:1804.06893  (2018)

\bibitem{zhu2017unpaired}
Zhu, J.Y., Park, T., Isola, P., Efros, A.A.: Unpaired image-to-image
  translation using cycle-consistent adversarial networks. In: Proceedings of
  the IEEE international conference on computer vision. pp. 2223--2232 (2017)

\end{thebibliography}

\end{document}